\documentclass[letterpaper]{article} % DO NOT CHANGE THIS
\usepackage[draft]{aaai2026}  % DO NOT CHANGE THIS
\usepackage{times}  % DO NOT CHANGE THIS
\usepackage{helvet}  % DO NOT CHANGE THIS
\usepackage{courier}  % DO NOT CHANGE THIS
\usepackage[hyphens]{url}  % DO NOT CHANGE THIS
\usepackage{graphicx} % DO NOT CHANGE THIS
\usepackage{natbib}  % DO NOT CHANGE THIS AND DO NOT ADD ANY OPTIONS TO IT
\usepackage{caption} % DO NOT CHANGE THIS AND DO NOT ADD ANY OPTIONS TO IT
\usepackage{algorithm}
\usepackage{algorithmic}
\usepackage{subcaption}

\usepackage{newfloat}
\usepackage{listings}
\usepackage{multirow}

\usepackage{booktabs} % for professional tables

\usepackage{csquotes}

\usepackage{siunitx}

\DeclareCaptionStyle{ruled}{labelfont=normalfont,labelsep=colon,strut=off} % DO NOT CHANGE THIS
\floatstyle{ruled}
\newfloat{listing}{tb}{lst}{}
\floatname{listing}{Listing}
\usepackage{amsmath}
\usepackage{amssymb}
\usepackage{mathtools}
\usepackage{amsthm}
\usepackage{bm}
\usepackage{multirow}

\usepackage{booktabs} % for professional tables

\usepackage[capitalize,noabbrev]{cleveref} % for better tables
\usepackage{csquotes}

\usepackage{siunitx}
\newrobustcmd\best{\DeclareFontSeriesDefault[rm]{bf}{b}\bfseries}

\title{rETF-semiSL: Semi-Supervised Learning for Neural Collapse in Temporal Data}

\author {
  Yuhan Xie\textsuperscript{\rm 1},
  William Cappelletti\textsuperscript{\rm 2},
  Mahsa Shoaran\textsuperscript{\rm 1},
  Pascal Frossard\textsuperscript{\rm 2}
}
\affiliations {
  \textsuperscript{\rm 1}INL, EPFL, Switzerland\\
  \textsuperscript{\rm 2}LTS4, EPFL, Switzerland\\
  yuhan.xie@epfl.ch
}

\begin{document}

\nocopyright
\maketitle

\begin{abstract}
Deep neural networks for time series must capture complex temporal patterns, to effectively represent dynamic data. Self- and semi-supervised learning methods show promising results in pre-training large models, which---when fine-tuned for classification---often outperform their counterparts trained from scratch. Still, the choice of pretext training tasks is often heuristic and their transferability to downstream classification is not granted, thus we propose a novel semi-supervised pre-training strategy to enforce latent representations that satisfy the Neural Collapse phenomenon observed in optimally trained neural classifiers. We use a rotational equiangular tight frame--classifier and pseudo-labeling to pre-train deep encoders with few labeled samples.
Furthermore, to effectively capture temporal dynamics while enforcing embedding separability, we integrate generative pretext tasks with our method, and we define a novel sequential augmentation strategy.
We show that our method significantly outperforms previous pretext tasks when applied to LSTMs, transformers, and state-space models on three multivariate time series classification datasets.
These results highlight the benefit of aligning pre-training objectives with theoretically grounded embedding geometry.
\end{abstract}

\section{Introduction}\label{intro}

Time series emerge across a wide range of applications where systems are monitored over time, and they represent successive observations from specific sources, which describe the state of the system.
For instance, wearable sensors such as smartwatch accelerometers can identify human activities, or electrical brain signals recorded via electroencephalograms might contain epileptic seizures.
To effectively classify this data, models must capture temporal dynamics as well as relationships between different variables, in multivariate settings.
Thus, deep learning models have been successfully applied to time series classification \citep{ismailfawazDeepLearningTime2019, mohammadifoumaniDeepLearningTime2024}, thanks to their ability to capture temporal patterns with mechanisms such as recursion or attention.
In particular, a popular approach is representation learning, that uses encoder-decoder architectures, where the encoding block extracts latent embeddings that are passed to a decoder layer, which is often a linear classifier.
However, deep models yet require large amounts of data to fully express their potential, and time series are often only sparsely labeled. For instance, in medical settings, patients are monitored over long term periods with symptoms appearing in rare and short events.

To leverage large time series datasets \cite{zhang2024selfsupervised} and compensate for scarce labels in specific tasks, strategies from semi- and self-supervised learning (SemiSL and SSL resp.) are commonly used.
Those methods define a \emph{pretext task} to capture inherent properties of the data at hand, and pre-train the encoder block with a task-specific \emph{pretext head}.
Then, the encoder and the original classifier are fine-tuned on a specific \emph{downstream} classification task.
SemiSL methods rely on at least part of the training labels from the downstream task to define the learning objective, and use different strategies to enforce generalization to unlabeled, and unseen data.
SSL, on the other hand, is more flexible, as it design pretext tasks on inherent properties of the data independently from downstream classification, but does not grant transferability of the learned embeddings to the latter.

To address this, we design a novel SemiSL task to learn from sparsely labeled data, which we further specialize for temporal data with SSL techniques.
Our task enforces \emph{Neural Collapse} (NC) \citep{papyanPrevalenceNeuralCollapse2020}, a behaviour observed in the final phases of deep neural networks training, where the embeddings of the last layer of different models converge, in the embedding space, to label-based clusters whose centers describe a simplex equiangular tight frame (ETF).
\citet{galanti2022role} demonstrate that such a distribution helps to generalize to new samples or even new classes.

Our novel framework, which we call \emph{rETF-semiSL}, pre-trains deep encoders on a few labeled samples using a pseudo-labeling strategy and a novel pretext head---consisting of a learnable feature-space rotation, a fixed classifier, and a specialized loss---which builds upon recent works \citep{hu2024neural, peifeng2023feature} that enforce Neural Collapse.
Furthermore, to promote invariance to temporal noise in the encoder, we combine rETF-semiSL with generative SSL and with a novel time series--specific augmentation method, which we call \emph{forward mixing}.
We use rETF-semiSL to pre-train common encoders architectures---such as RNNs, transformers, and state-space models---on three multivariate time-series datasets, and show that our ETF-based semi-supervised pre-training framework consistently outperforms established SSL and supervised methods, with an average relative improvement of downstream classification of 12\%.
Furthermore, our method has a lower computational complexity than other considered approaches, and grants faster convergence in finetuning.

We summarize our contributions as follows
\begin{enumerate}
  \item We propose \emph{rETF-semiSL}, a novel semi-supervised pre-training framework to enforce Neural Collapse in latent embeddings of deep encoders;
  \item We present a specialization of our method for temporal data, integrating generative SSL tasks as noise learning methods, and proposing a novel time series augmentation strategy;
  \item We show with extensive experiments that our framework consistently outperforms previous methods from both SemiSL and SSL literature, highly boosting the classification capabilities of modern deep models.
\end{enumerate}

\subsubsection{Paper overview.}
In \cref{sec:methods} we present the task description and different building blocks of our original semi-supervised learning method.
In \cref{sec: rel works} we introduce related studies about time series pretext tasks, transferability and Neural Collapse.
Then in \cref{sec:experiments} we present our experiments on time series classification, with in depth analysis and ablation studies.

\section{Related Work}\label{sec: rel works}

We present prior works about enforcing Neural Collapse (NC) to improve classification performance, and explain the novelty of our method, which is the first to consider NC within a pre-training objective.
Then, we present popular pretext tasks from the SSL and SemiSL literature, against which we compare in the experimental section.
Finally, we focus on data augmentation techniques, commonly applied in previous methods to enforce inherent properties of the data to the embeddings, which highlight the relevance of our novel strategies for temporal data.

%-------------------------------------------------------------------------------
% NC, ETF, etc.

Motivated by the optimal structure of NC, multiple works \citep{liu2023inducing,xie2023neural} propose classification models using randomly initialized equiangular tight frames (ETF) as fixed classifiers, and they observed significant improvements in classification performances and faster convergence on imbalanced datasets.
In order to train the encoder with the ETF, \citet{peifeng2023feature} uses the simple Cross Entropy loss,  and \citet{yang2022inducing} proposes the Dot Regression loss, observing faster convergence and robustness against data imbalance. We propose a center loss to enhance the clustering of embeddings.
% \citet{kimFixedNonnegativeOrthogonal2023} introduce a fixed non-negative orthogonal classifier to achieve zero-mean Neural Collapse.
Getting one step further, \citet{gao2024distribution} optimize simultaneously the classifier and encoder towards the ETF structure.
\citet{hu2024neural} propose a semi-supervised learning method that pre-trains the encoder through a fixed ETF classifier and achieves good performance on image classification tasks.
We extend this framework, first by integrating rotational feature space \citep{peifeng2023feature} and a center loss into the ETF classifier model; then, designing a specific pseudo-labeling algorithm; and finally by using time-series-specific augmentation methods.

%-------------------------------------------------------------------------------
% SSL and SemiSL

In SSL and SemiSL, contrastive and generative models are two popular approaches.
Contrastive-based methods \citep{jaiswal2020survey} are widely used in computer vision and are designed to learn instance discrimination \citep{wu2018unsupervisedfeaturelearningnonparametric} in the embedding space.
Recent works by \citet{schneider2022detecting}, \citet{pranavan2022contrastive} and \citet{Eldele_2023} show that contrastive learning methods are effective in learning temporal representations, and they often improve downstream classification performance.
These methods define \emph{positive} and \emph{negative pairs}, with the former consisting of \enquote{similar} inputs, while the latter of other data. The encoder is trained to minimize the distance between embeddings of positive pairs, and push apart negative ones.
A prominent SSL example is \emph{SimCLR} \citep{chen2020simple}, where positive examples are small perturbations of a sample in the dataset, and negative ones are different random samples.
On the other hand, contrastive SemiSL methods exploit label information to define pairs and enforce the separability of classes in the latent space.
\citep{khosla2021supervised} propose \emph{supervised contrastive learning} (SupCon) which further boosts classification performance.
Recent work \citep{graf2021dissecting, xue2023features} show that supervised contrastive learning optimizes features towards NC, which partially aligns it to the objective of our framework.
We claim that what is for SupCon a byproduct, is the property that enhances its performances.

Generative-based methods \citep{zhang2022survey} learn representations of time series by training the encoder and pretext decoder to either predict future values, as in autoregressive forecasting \citep{seq2seq}, or hold-out parts of the sequence, as in masked time series reconstruction \citep{li2023ti}.
These SSL methods are widely adopted with the assumption
that they can effectively learn task-agnostic features, by leveraging the temporal consistency of time series.
Still, how pre-trained models adapt to specific downstream tasks, or 
\emph{transferability}, is an active research topic in transfer learning (TL) literature \citep{zhuangComprehensiveSurveyTransfer2021}, and no prior work studied the transfer from generative SSL on temporal data to time series classification tasks.
Even though multiple transferability metrics have been proposed \citep{jiang2022transferability}, the choice these methods as pretext tasks is still heuristic, and improvement in downstream performances against a model trained from scratch is not guaranteed, as TL consider each task equally, while in SemiSL and SSL the pretext task is subordinate to the downstream one.
Still, only generative tasks directly leverage the temporal structure of time series by reconstructing or predicting parts of the sequence, while contrastive methods often rely on augmentations that may disrupt temporal dependencies.
For this reason, we propose a specific temporal augmentation for our SemiSL method, and we study joint learning with autoregressive tasks.

%-------------------------------------------------------------------------------
% Augmentation

Many SSL and SemiSL methods employ data augmentation techniques, including time-domain, frequency-domain methods, and learning-based augmentations \cite{wen2020time}. These methods aim to sample from an estimated training distribution $P$($\tilde{x}$), where $\tilde{x}$ denotes augmented samples. A natural way to estimate $P$($x$) is by randomly mixing existing samples \cite{jin2024survey}, which has been applied to time series in prior work \citep{wickstrom2022mixing, demirel2023finding, tangSelfSupervisedGraphNeural2021,Eldele_2023}. In contrast, we propose a new mixing strategy that linearly interpolates between adjacent time steps, based on the assumption that intermediate representations can be estimated by segment-wise interpolation.
\section{Methods}\label{sec:methods}

\begin{figure}[t]
    \centering

    \begin{subfigure}{\linewidth}
        \centering
        \includegraphics[width=0.9\linewidth]{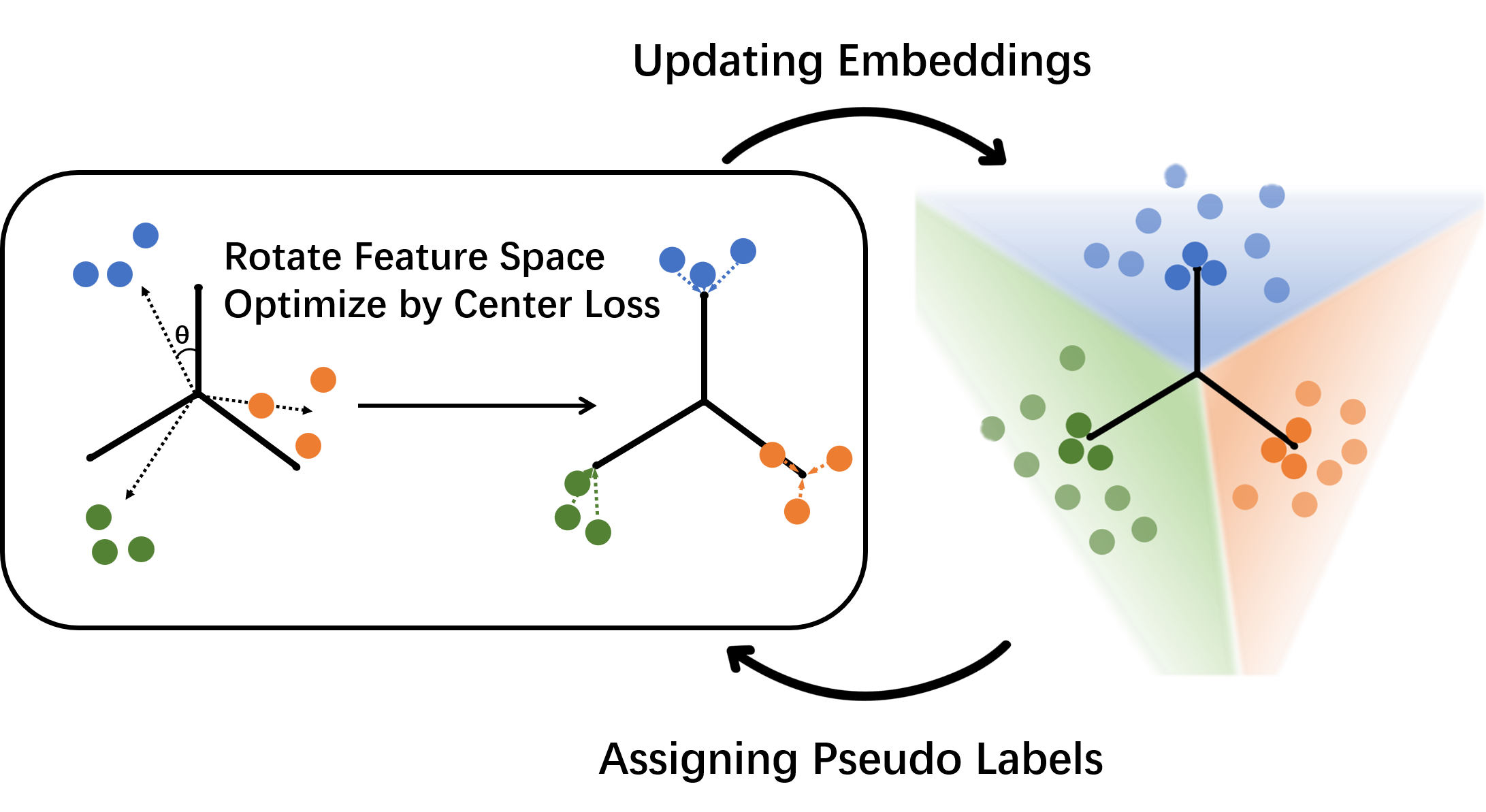}
        \caption{Illustration of our semi-supervised training process. Our alternating update process consists of two parts: On the left, we leverage a rotational matrix and the DR Loss to optimize embedding space towards Neural Collapse; On the right part, deep colored points represent labeled samples, and light colored ones are samples assigned pseudo labels. The colored regions illustrate the pseudo labelling rule based on the ETF structure.}
        \label{fig:main_a}
    \end{subfigure}
    
    \vspace{0.5em}
    
    \begin{subfigure}{\linewidth}
        \centering
        \includegraphics[width=0.8\linewidth]{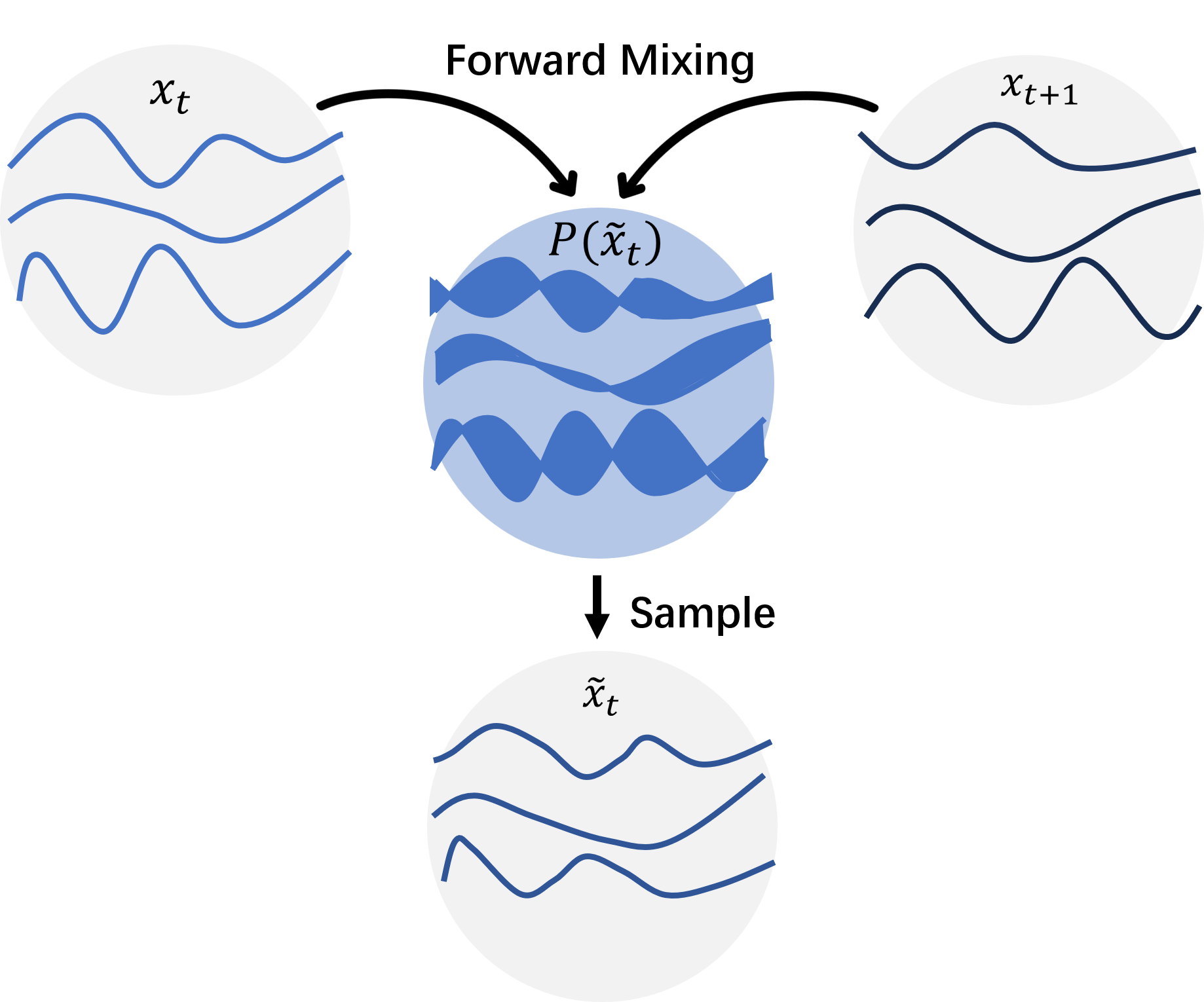}
        \caption{Illustration of the forward mixing approach. For time step t, $x_t$ is augmented by drawing a sample from $P(\tilde{x}_t)$, which is generated by interpolating between $x_t$ and $x_{t+1}$ with a certain proportion.}
        \label{fig:main_b}
    \end{subfigure}

    \caption{Illustration of our proposed method.}
    \label{fig:main}
\end{figure}

We define our novel ETF-based semi-supervised learning framework for time series classification. We begin by leveraging a neural network architecture consisting Equiangular Tight Frame (ETF) classifier and learnable rotational matrix; then we present a loss function for feature clustering, and a novel semi-supervised learning strategy.
Finally, we specialize our model to time series data by introducing a sequential augmentation method.

Given a dataset of $n$ time series samples $\{(X_i, y_i)\}_{i=1}^n$, where $X_i \in \mathbb{R}^{T \times N}$ denotes a time series of length $T$ with $N$ variables, and $y_i \in \{1, \dots, K\}$ is the corresponding class label, our goal is to learn a function $h_\theta: \mathbb{R}^{T \times N} \rightarrow \{1, \dots, K\}$ parameterized by $\theta$ that accurately predicts the class label for unseen samples.
% When $N > 1$, it becomes a multivariate time series, which is harder to deal with, so we focus on multivariate time series in experiments.

\subsection{Problem Formulation}

The model $h_\theta$ is composed of an encoder $f_\theta: \mathbb{R}^{T \times N} \rightarrow \mathbb{R}^d$ that maps the input time series to a $d$-dimensional latent representation, and a classifier $g_\phi: \mathbb{R}^d \rightarrow \mathbb{R}^K$ that maps the latent representation to class logits. The overall prediction is given by $h_\theta(X) = g_\phi(f_\theta(X))$.

The parameters $\theta$ and $\phi$ are optimized to minimize a classification loss $\mathcal{L}(h_\theta(X_i), y_i)$ over the labeled dataset. We observe time series $X_i \in \mathbb R^{T \times N}$ with associated class labels $y_i \in 1, \dots, K$ from a distribution~$\mathcal D$, and we aim to find the optimal parameters $\theta^*$ for a neural network $h_\theta$ that minimize a classification loss~${ \mathcal L \left( h_\theta (X_i), y_i \right) }$.
More precisely, we consider a setting were $h = g \circ f$, with $f$ being an encoder network that maps $X$ to a latent space, and $g$ being a linear layer, i.e. a matrix, that maps latent embeddings $f(X)$ to class logits.

The performance of the model $h = g \circ f$ is highly dependent on the quality of the embeddings of its encoder $f$.
Self- and semi-supervised learning methods are typically used to find initial parameters for the encoder, where no or only partial labels are accessible, that provide a better starting point than random initialization.
They pre-train $f$ on different \emph{pretext} tasks, in conjunction with specific \emph{task heads}~$g_{pre}$, then fine-tune it for the classification loss $\mathcal L$ with the classifier head $g$.

%%%%%%%%%%%%%%%%%%%%%%%%%%%%%%%%%%%%%%%%%%%%%%%%%%%%%%%%%%%%%%%%%%%%%%%%%%%%%%%%

\subsection{ETF-based Semi-supervised learning}\label{ssec:etf-semisl}

We design an ETF-based semi-supervised learning--pretext task, which we call \emph{rETF-semiSL}.
It combines an ETF classification layer with a learnable rotation and a specific loss to enforce Neural Collapse in latent representations.
We propose a novel alternating-update learning strategy to leverage large datasets with sparse labels.

\subsubsection{Neural Collapse.} \citet{papyanPrevalenceNeuralCollapse2020} observe that during the terminal phase of deep neural networks training the embeddings of the last layer of different models exhibit \emph{Neural Collapse}, which consists of four properties:
\begin{itemize}
    \item (NC1) Embedding variance within each class converges to zero;
    \item (NC2) Class mean vectors relative to the global mean in the embedding spaces converge to equally spaced vertices of a simplex;
    \item (NC3) Class mean vectors converge to the class vector, i.e. column, of the classifier;
    \item (NC4) The readout layer is equivalent to a nearest neighbour classification in the latent space.
\end{itemize}

\subsubsection{ETF classifier.}\label{ssec:etf-classifier}

\citet{yang2022inducing} propose an encoder-decoder model, whose classifier block consists of a simplex equiangular tight frame (ETF) and an orthogonal matrix.
Let $K$ be the number classes, and the encoder $f \mapsto \mathbb R^d$, we define the ETF classifier as
\begin{equation} \label{eq:ETF}
\mathbf{W}=\sqrt{\frac{K}{K-1}} \mathbf{U}\left(\mathbf{I}_K-\frac{1}{K} \mathbf{1}_K \mathbf{1}_K^T\right),
\end{equation}
where $\mathbf{I}_K \in \mathbb{R}^{K \times K}$ is an identity matrix, $\mathbf{1}_K \in \mathbb{R}^{K}$ is an all-one vector, and
$\mathbf{U} \in \mathbb{R}^{d \times K}$ is a orthogonal matrix that satisfies $\mathbf{U}^T\mathbf{U} = \mathbf{I}_K$.
The role of $\mathbf{U}$ is to project the $d$-dimensional embedding vector to the $K$-dimensional space of class logits.

By definition, each column $\mathbf{w}_i$ of $\mathbf{W}$ lies on the $d$-dimensional unit sphere, and the angle between any two $\mathbf{w}_i, \mathbf{w}_i$ is equal and maximized, which means $\mathbf{w}_i^T \mathbf{w}_j=-\frac{1}{K-1}, i \neq j$. These two features ensure that the classifier will not be biased by the dataset imbalance, and they force latent representations to converge to Neural Collapse.

\subsubsection{Rotated Balanced Representations.}

To reduce the feature learning difficulty, we enable~$\mathbf{W}$ to rotate while maintaining the ETF property, using the rotated balanced learning framework from \citet{peifeng2023feature}. We introduce a learnable rotation matrix~$\mathbf R \in \mathbb{R}^{d \times d}$, with $\mathbf R^\top \mathbf R = \mathbf{I}_d$, between the embedding block~$f$ and the classifier~$\mathbf{W}$, resulting in $\mathbf{W^r} = \mathbf{R W}$.
In order to optimize $\mathbf R$ under its constraint with SGD, we use a trivialization tool provided by \citet{10.5555/3454287.3455108}.

\subsubsection{Center Loss.}

To further improve the clustering of embeddings under an ETF classifier, we introduce a modified center loss.
Unlike the version proposed by \citet{wen2016discriminative}, we add a direction term to combine with the distance term, and control the weight by $\alpha$:
\begin{equation}
\mathcal{L}_{Center}\left(\mathbf{h}, \mathbf{W^r}\right)
    = 1 - \frac{\mathbf{h}^{T} \mathbf{w^r}_c}{\|\mathbf{h}\| \|\mathbf{w^r}_c\|} + \alpha \|\mathbf{h} - \mathbf{w^r}_c\|,
\end{equation}
where $\mathbf{h} = f(X)$ is the embedding vector, and $\mathbf{w^r}_c$ represents corresponding class vector. Compared to Cross Entropy (CE) loss \citep{peifeng2023feature} and Dot Regression (DR) loss \citep{yang2022inducing}, our center loss optimizes directly toward an ETF structure because it is minimized only when $\mathbf{h}$ and $\mathbf{w^r}_c$ completely overlap.

%-------------------------------------------------------------------------------

\subsubsection{Semi-supervision by Alternating Updates.}

\begin{algorithm}[t]
\caption{SemiSL alterating updates of rETF classifier}
\label{alg:semi-supervised-etf}
\begin{algorithmic}[1]
\REQUIRE Labeled dataset $\mathcal{D}_L$, unlabeled dataset $\mathcal{D}_U$, encoder $f$, ETF classifier $\mathbf{W^r}$

\STATE \textbf{Step 1: Supervised Initialization}
\STATE Initialize encoder $f$ using labeled dataset $\mathcal{D}_L$

\WHILE{not converged }
    \STATE \textbf{Step 2: Pseudo-Labeling}
    \STATE Use the current encoder $f$ to infer features on $\mathcal{D}_U$
    \STATE Use the ETF classifier to compute softmax probabilities and assign pseudo-labels to $\mathcal{D}_U$

    \STATE \textbf{Step 3: Semi-supervised Representation Learning}
    \STATE Use labeled data $\mathcal{D}_L$ and pseudo-labeled data $\mathcal{D}_U$ to update encoder $f$
\ENDWHILE
\end{algorithmic}
\end{algorithm}

To enable the encoder to learn a representation closer to Neural Collapse with limited labeled samples, we design \cref{alg:semi-supervised-etf}.
In Step 1, we initialize the model by training with the labeled subset~$\mathcal{D}_L$. Given a randomly initialized classifier~$\mathbf{W^r}$ and sample~$X_l$ from~$\mathcal{D}_L$, the model optimizes the encoder~$f$ and the rotational matrix~$\mathbf R$ by minimizing $\mathcal{L}_{Center}\left(f(X_l), \mathbf{W^r}\right)$.
In Step~2, we compute the classification logits~$L$ for each unlabeled sample~$X_u$ as $L = \mathbf{W^r}^T f(x_u), L \in \mathbf{R}^K$, and assign pseudo label $\tilde y_u = \operatorname{argmax} L$. In Step 3, the model is further optimized on $\mathcal L_{Center}$ using both $\mathcal{D}_L$, and $\mathcal{D}_U$ with pseudo labels.
After initialization, we iteratively perform Step~2 and Step~3 in each epoch.

%%%%%%%%%%%%%%%%%%%%%%%%%%%%%%%%%%%%%%%%%%%%%%%%%%%%%%%%%%%%%%%%%%%%%%%%%%%%%%%%

\subsection{Learning Time Series Invariance}

To leverage properties specific to time series, we focus on learning strategies and augmentations that explicitly exploit the sequential structure of the data. First, we describe how generative self-supervised learning (SSL) tasks---such as sequence reconstruction and forecasting---act as noise processes during training, and we propose to use them as auxiliary tasks to our \emph{rETF-semiSL}.
Then, we introduce a time series-specific data augmentation, namely \emph{forward mixing}, which generates augmented samples by interpolating between consecutive time steps.

%---------------------------------------------------------------------

\subsubsection{Generative SSL Tasks as Noise Processes.}\label{ssec:mse-decomposition}

We now examine the MSE loss, commonly used in autoregressive and forecasting tasks, to explain its functions and limitations as a pretext loss for classification.
We study the average MSE loss by the marginal probability~$P(i)$ of each class, which gives
{\small\begin{equation}
\mathbb{E}%_{\mathcal{X}^t, \mathcal{Y}^t}
  \left\|g \circ f\left(\mathcal X^t\right)- \mathcal Y^t\right\|_2^2
=\sum_{i=1}^k \mathbb{E}%_{\mathcal{X}_i^t, \mathcal{Y}_i^t}
  \left\|g \circ f\left(\mathcal X_i^t\right)- \mathcal Y_i^t\right\|_2^2 P(i),
\end{equation}}
where $\mathcal{X}_i^{t}$ and $\mathcal{Y}_i^{t}$ are the distribution of the input signal and the target signal at time t for class i, and $g \circ f(x)$ is the input signal passed through encoder $f$ and decoder $g$.
In reconstruction tasks $\mathcal{Y}_i^t = \mathcal{X}_i^t$, while for $k$-steps ahead prediction $\mathcal{Y}_i^t = \mathcal{X}_i^{t+k}, k \geq 0$; we will denote the distribution of $g \circ f(\mathcal X_i^t)$ by $\mathcal{\hat{Y}}_i^t$. In supplementary material we detail the decomposition of the MSE Loss:
{\small\begin{equation}\label{loss decomp}
\begin{aligned}
\operatorname{MSE} &= \mathcal{L}_1 + \mathcal{L}_2 + \mathcal{L}_3 + \sum_{i=1}^k \operatorname{Var}\left(\mathcal{Y}_i^t\right) P(i) ,\\
\mathcal{L}_1 &=\sum_{i=1}^k \operatorname{Var}_{g \circ f}\left(\mathcal{X}_i^t\right)P(i),\\
\mathcal{L}_2 &= \sum_{i=1}^k\left\|\mu_{g \circ f}\left(\mathcal{X}_i^t\right)-\mu\left(\mathcal{Y}_i^t\right)\right\|_2^2P(i),\\
\mathcal{L}_3 &= -2 \sum_{i=1}^k \operatorname{Tr}\Big(\operatorname{Cov}(\mathcal{\hat{Y}}_i^t, \mathcal{Y}_i^t)\Big)P(i).\\
\end{aligned}
\end{equation}}\\
The term $\operatorname{Var}\left(\mathcal{Y}_i\right)$ represents the variance of the supervision signal, which we cannot optimize.
$\mathcal{L}_1$ represents the variance of the predicted signal, $\mathcal{L}_2$ represents the distance between the mean of the predicted distribution and the mean of its target class distribution, and $\mathcal{L}_3$ represents the element-wise covariance between the input and output signals, measuring the shape similarity between two signals. We observe that minimizing the first and second terms may introduce some degree of separability in the embedding space as well as reducing the variance of the embeddings, but this is constrained by the original data distribution and the encoder's contribution to the loss.
In the supplementary material, we conduct experiments on synthetic data to verify this claim.

%---------------------------------------------------------------------

\subsubsection{Forward-Mixing Augmentation}\label{gen}

 While generative tasks can help the model learn time series invariance, their optimization objectives are not always aligned with the geometry induced by Neural Collapse.
To address this limitation, we propose to enforce this invariance in the model directly through augmenting data.
Compared to conventional augmentation methods \citep{zhang2022self}, which are generally agnostic to temporal structure and input distribution, our approach generates augmented samples with a sequential paradigm.
Specifically, given an input time series $x_t \in \mathbf{R}^{T \times N}$, we generate an augmented version
\begin{equation}
    \tilde{x}_t = x_t + \sigma * (x_{t+1} - x_t),
\end{equation}
where $\sigma \sim \mathcal{U}(0, p)$ is a noise coefficient sampled from a uniform distribution, where $p$ is an hyperparameter.
This technique, which we call \textbf{forward mixing}, creates smooth perturbations based on the local temporal interpolation.
Compared to adding random noise globally, forward mixing respects the \textit{variable-wise dynamics} of the signal, making it more aligned with the temporal structure of time series.

\section{Experiments}\label{sec:experiments}

\begin{table*}[t]
\centering
\scriptsize

\label{tab:class-multi-model}
\setlength{\tabcolsep}{3pt}
\begin{tabular}{l
    r@{ \textpm }l r@{ \textpm }l r@{ \textpm }l
    r@{ \textpm }l r@{ \textpm }l r@{ \textpm }l
    r@{ \textpm }l r@{ \textpm }l r@{ \textpm }l
    r@{ \textpm }l r@{ \textpm }l r@{ \textpm }l}
\toprule
\scshape Method &
\multicolumn{6}{c}{\scshape LSTM} &
\multicolumn{6}{c}{\scshape iTransformer} &
\multicolumn{6}{c}{\scshape TimesNet} &
\multicolumn{6}{c}{\scshape Mamba} \\
\cmidrule(lr){2-7}
\cmidrule(lr){8-13}
\cmidrule(lr){14-19}
\cmidrule(lr){20-25}
\scshape Dataset & \multicolumn{2}{c}{HAR} & \multicolumn{2}{c}{Epilepsy} & \multicolumn{2}{c}{Heartbeat} 
& \multicolumn{2}{c}{HAR} & \multicolumn{2}{c}{Epilepsy} & \multicolumn{2}{c}{Heartbeat}
& \multicolumn{2}{c}{HAR} & \multicolumn{2}{c}{Epilepsy} & \multicolumn{2}{c}{Heartbeat}
& \multicolumn{2}{c}{HAR} & \multicolumn{2}{c}{Epilepsy} & \multicolumn{2}{c}{Heartbeat} \\
\midrule

\itshape Base & 
20.5 & 1.06 & 19.6 & 3.99 & 51.4 & 4.12 & 
51.3 & 1.49 & 38.0 & 2.62 & \underline{68.9} & \underline{2.47} &
\underline{80.9} & \underline{0.48} & \underline{64.9} & \underline{2.99} & 54.4 & 1.91 &
64.9 & 3.40 & 16.5 & 5.06 & 48.2 & 7.87 \\

\midrule

\itshape + Auto & 
46.9 & 0.52 & 37.4 & 3.27 & 53.8 & 9.13 &
47.8 & 0.41 & \underline{46.2} & \underline{4.12} & 45.7 & 9.66 &
80.7 & 1.55 & 58.3 & 1.87 & 57.3 & 1.17 &
62.2 & 1.54 & 14.5 & 6.11 & \underline{53.7} & \underline{2.60} \\

\itshape + Recon & 
47.1 & 0.09 & 33.5 & 6.55 & 54.5 & 12.4 &
51.2 & 0.75 & 37.8 & 4.82 & 55.3 & 10.6 &
80.4 & 1.17 & 55.5 & 3.81 & 57.8 & 1.93 &
34.2 & 2.46 & 13.9 & 5.77 & 51.3 & 6.25 \\

\itshape + Con & 
58.3 & 0.93 & 40.1 & 3.57 & \bfseries 66.2 & \bfseries 3.18 &
\underline{61.8} & \underline{0.20} & 38.7 & 2.09 & 56.8 & 9.38 &
76.9 & 1.89 & 57.2 & 7.02 & \underline{59.3} & \underline{0.95} &
67.3 & 0.29 & 27.4 & 16.7 & 40.2 & 3.48 \\

\itshape + SupCon & 
\underline{78.4} & \underline{0.09} & \underline{40.5} & \underline{2.11} & 61.5 & 0.84 &
56.3 & 2.37 & 41.9 & 4.70 & 46.7 & 17.7 &
66.5 & 5.52 & 48.4 & 13.8 & 55.4 & 5.61 &
\underline{91.6} & \underline{0.02} & \underline{77.1} & 24.4 & 47.7 & 2.94 \\

\underline{\itshape + rETF-semiSL} & 
\bfseries 86.9 & \bfseries 0.23 & \bfseries 48.9 & \bfseries 1.89 & \underline{64.0} & \underline{2.42} &
\bfseries 83.9 & \bfseries 0.02 & \bfseries 48.2 & \bfseries 1.79 & \bfseries 69.9 & \bfseries 0.72 &
\bfseries 90.1 & \bfseries 0.36 & \bfseries 86.2 & \bfseries 0.54 & \bfseries 59.5 & \bfseries 1.37 &
\bfseries 92.4 & \bfseries 0.66 & \bfseries 89.7 & \bfseries 2.10 & \bfseries 60.8 & \bfseries 2.73 \\

\bottomrule
\end{tabular}

\caption{Fine-tuned classification results with different pretext tasks (\underline{ours} are underlined), on HAR, Epilepsy, and Heartbeat datasets across four models. We use F1 score for HAR and Epilepsy, and AUROC for Heartbeat. All scores are shown in percentage (\%) and rounded to three significant digits. For all tasks, higher metrics are better, \textbf{best scores} are bold and \underline{second best} are underlined. We report the average and STD test values obtained across 5 random initializations of each model.}

\end{table*}

We compare rETF-semiSL against established methods for pre-training deep neural networks in multivariate classification tasks. We present experiments on multiple backbone models and datasets, and investigate downstream task performance, transferability, and computational efficiency. We first compare rETF-semiSL with other supervised and self-supervised methods in \cref{classification etf}, and then we compare forward-mixing with generative joint training on rETF-semiSL in \cref{ssec:exp joint}.

\subsection{Experimental Settings}
To fairly compare different methods, we use a backbone model to train different pretext tasks, then fix the backbone model and fine-tune with one linear layer on the downstream task, which is called linear probing. The following pretext tasks are included in the experiment:
\begin{itemize}
  \item \textit{Reconstruction (Recon)}: reconstruct input signal with MSE loss;
  \item \textit{Autoregressive Prediction (Auto)}: predict 1-step ahead signal with MSE loss;
  \item \textit{Contrastive Learning (Con)}: create augmented samples by adding Gaussian noises. There is 1 positive sample and $2N-2$ negative samples for each sample in the batch if the batch size is $N$. We use the same loss function and MLP projector as SimCLR \cite{chen2020simple};
  \item \textit{Supervised Contrastive Learning (SupCon)}: same setting as \textit{Con}, except that \textit{SupCon} leverages label information to create positive and negative samples. For a batch of $N$ samples, there are $2K-1$ positive samples and $2N-2K$ negative samples for each sample, where the class size of the sample in the batch is $K$. The loss function and projector are aligned with the setting of \citet{khosla2021supervised}.

\end{itemize}

We also use the randomly initialized model \textit{Base} as the baseline. We use four different backbone models, LSTM \citep{hochreiterLongShortTermMemory1997}, iTransformer \citep{liuitransformer}, TimesNet \citep{wutimesnet} and Mamba \citep{gumamba}, which cover various scale and architecture. To ensure a comprehensive and balanced evaluation, datasets and models are selected based on their diversity in channel and class configurations, as well as model scale and architectural variety.
We use three time series classification datasets for the experiments, namely Epilepsy and Heartbeat from UEA datasets \citep{bagnall2018uea} and Human Activity Recognition (HAR) \citep{Anguita2013APD}.
For semi-supervised training, we sample 30\% data as a labeled subset for HAR, and 70\% for Epilepsy and Heartbeat. Subsets have the same stratification of labels as the training set. To ensure the algorithm is robust to different labeled subsets, we randomly initialize the seed for sampling for each test.
 Further details about experiment settings can be found in the supplementary material.

%-----------------------------------------------------------------

\subsection{Experiment Results}

\subsubsection{Downstream Task Performance} \label{classification etf}
\cref{tab:class-multi-model} shows the downstream classification results with different pretext methods. We see that our rETF-semiSL method achieves the best performance across the three datasets. The second best method Supervised Contrastive learning and self-supervised Contrastive learning.

These results demonstrate that our method achieves excellent performance across different datasets and exhibits strong compatibility with four different models.
We observe that variability of the autoregressive task, which proves useful with LSTM, but reduces the performance with the other three backbone models, is due to the overfitting on the inherent time series noise. More precisely, in supplementary material, we observe that the autoregressive loss is dominated by the minimization of the mean embeddings between the reconstructed and target signal distribution, which might be independent from labels.

%--------------------------------------------------------------------------------

\subsubsection{rETF-semiSL with Time Series Constraints}\label{ssec:exp joint}

\begin{table}[t]

\setlength{\tabcolsep}{3pt}
\centering

\label{tab:joint-auto2}

\resizebox{\linewidth}{!}{
\sisetup{
  detect-weight,
  mode=text,
  reset-math-version=false,
  table-format=1.2,
}
\begin{tabular}{l S S S S}
\toprule
\scshape Dataset &
\scshape { LSTM } &
\scshape { iTransformer } &
\scshape { TimesNet } &
\scshape { Mamba } \\
\midrule

\scshape HAR \textit{+Auto} &
{--} &
\best +3.09\%$^{*}$ &
{--} &
{--} \\

\scshape HAR \textit{+Noise} &
{--} &
+2.28\%$^{*}$ &
{--} &
\best +2.02\%$^{**}$ \\

\scshape HAR \textit{+Mix} &
{--} &
+2.79\%$^{*}$ &
{--} &
\best +2.02\%$^{**}$ \\

\midrule

\scshape Epilepsy \textit{+Auto} &
+6.49\%$^{*}$ &
+5.06\%$^{**}$ &
{--} &
{--} \\

\scshape Epilepsy \textit{+Noise} &
\best +9.97\%$^{**}$ &
+5.27\% &
+0.93\%$^{*}$ &
{--} \\

\scshape Epilepsy \textit{+Mix} &
+8.73\%$^{**}$ &
  \best  +11.4\%$^{**}$ &
\best +3.45\%$^{**}$ &
{--} \\

\midrule

\scshape Heart \textit{+Auto} &
{--} &
{--} &
{--} &
{--} \\

\scshape Heart \textit{+Noise} &
{--} &
{--} &
{--} &
{--} \\

\scshape Heart \textit{+Mix} &
\best +2.89\% &
{--} &
{--} &
{--} \\

\bottomrule
\end{tabular}
}
\caption{Downstream classification results of rETF-semiSL with two regularization methods: joint learning with autoregressive task ($\textit{+ Auto}$), normal augmentation with Gaussian Noise ($\textit{+Noise}$), and forward mixing augmentation ($\textit{+ Mix}$). `--' means the method doesn't show improvement compared to rETF-semiSL; otherwise, the average improvement percentage is reported. Best performance of each dataset with certain model is \textbf{bolded}. $^{*}$ represents p-value $<$ 0.05, and $^{**}$ represents p-value $<$ 0.01.}

\end{table}

In this section, we compare the performances of forward mixing augmentation and generative auxiliary pretext tasks.
These methods enforce time series constraints in the encoder, to efficiently remove the inherent noise in the temporal process, and in the data distribution.
% \subsubsection{Forward Mixing Improves Downstream Performance}
For generative method, We define the following joint loss for rETF-semiSL and autoregressive prediction :
\begin{equation}
\label{joint loss}
\mathcal{L}_{joint} = \mathcal{L}_{Center} + \frac{\alpha | \mathcal{L}_{Center} |}{|\mathcal{L}_{auto}|} \mathcal{L}_{auto},
\end{equation}
where $\mathcal{L}_{Center}$ is the center loss, and $\mathcal{L}_{auto}$ is the MSE loss. We first scale $\mathcal{L}_{auto}$ to the same value as $\mathcal{L}_{Center}$, and then rescale it with $\alpha$. In practice, we select $\alpha$ by grid search, by comparing finetuning performance on the prediction task. For forward mixing, we generate noisy input $\tilde{x_t}$ as described in \cref{ssec:mse-decomposition} when training rETF-semiSL. In comparison, we use augmentation with gaussian noise as a baseline. Instead of using a residual part as the noise, we use gaussian noise $\mathcal{N}(0,\sigma^2)$, where variance is sampled from the $\mathcal{U}(\sigma^2_{x_t}, \sigma^2_{x_{t+1}})$, also with a random scale from a uniform distribution $\mathcal{U}(0,p)$.

\cref{tab:joint-auto2} presents the downstream classification results. We compute the significance of performance between using single rETF-semiSL and rETF-semiSL with these constraints, and report the p-value from Welch's t-test. Overall, forward mixing tends to improve the performance of rETF-semiSL more consistently.
Interestingly, adding Gaussian noise slightly boost performances, even though less than our proposed method. This shows that replacing Gaussian noise with the noise conditioned by the input distribution actually benefits the representation. In contrast, joint learning with the autoregressive task yields performance gains only on specific models and datasets.  The results suggest that, to effectively leverage the temporal dependency from $x_t$ to $x_{t+1}$ as a useful prior for classification, generating interpolated samples is a more effective approach.

%------------------------------------------------------------------

\subsubsection{Ablation Study}\label{ssec:exp-ablation}

We perform an ablation study on rotational matrix $\mathbf{R}$, center Loss $\mathcal{L}_{Center}$ and our pseudo-labeling method to show that they are all necessary components for good representations. We use \textit{\{$\mathbf{R}$, $\mathcal{L}_{Center}$\} } as a baseline and compare it with the model with \textit{\{$\mathbf{R}$, $\mathcal{L}_{CE}$\}}, \textit{\{$\mathbf{R}$, $\mathcal{L}_{DR}$\}} or \textit{\{$\mathcal{L}_{Center}$\}}. For pseudo-labeling, we compare our method with Nearest-neighbour Labeling (\textit{NN Labeling}) by using $L_2$ distance to assign labels, or \textit{K-means Labeling} that is proposed by \cite{hu2024neural}. According to \cref{ablation-result}, both $\mathbf{R}$ and $\mathcal{L}_{Center}$ are important for the model to get stable performance across 3 datasets, and $L_{Center}$ outperforms $ \mathcal{L}_{CE}$, $\mathcal{L}_{DR}$. Our simple pesudo-labeling method also achieves better results than \textit{K-means Labeling}, without the need to recompute the cluster centers on the dataset.

\begin{table}[t]

\centering

\label{ablation-result}
\setlength{\tabcolsep}{1.1pt}

\begin{tabular}{lcccc}
\toprule
\scshape Conditions & \scshape HAR & \scshape Epilepsy & \scshape Heartbeat & \scshape Avg \\
\midrule
\itshape \{$\mathbf{R}$, $\mathcal{L}_{CE}$\}     & +1.13\% & -1.90\% & -8.33\% & -2.59\% \\
\itshape \{$\mathbf{R}$, $\mathcal{L}_{DR}$\}     & -4.98\% & +4.10\% & -6.29\% & -2.59\% \\
\itshape $\{\mathcal{L}_{Center}\}$              & -2.49\% & -0.78\% & -1.89\% & -1.77\% \\
\midrule
\itshape NN Labeling                            & -16.8\% & -8.78\% & -4.72\% & -10.8\% \\
\itshape K-means Labeling                                      & -14.3\% & -1.32\% & -4.24\% & -7.36\% \\
\bottomrule
\end{tabular}

\caption{Ablation experiment results on HAR, Epilepsy, and Heartbeat datasets. In the table, the relative performance change is reported across four models for each dataset. The average performance change of all datasets is reported in the last column.}

\end{table}

%---------------------------------------------------------------------

\subsubsection{Quantifying Pretext Task Representations}\label{ssec:exp-pretext-metrics}

In order to directly evaluate how effectively the latent representation from pretext tasks enhances downstream performance, we leverage three popular transferability metrics: NLEEP \citep{li2022ranking}, LogME \citep{you2021logme} and SFDA \citep{shao2022models}.
They quantify the probability $P(y|z)$ of predicting the downstream label~$y$ from the pre-trained embedding~$z$.
A higher metric value indicates the representation is easier to get higher accuracy in classifying the given labels.

From a different perspective, \citet{galanti2022role} introduce the class-distance normalized variance (CDNV) to estimate whether a representation satisfies Neural Collapse.
If the class average CDNV value approaches zero, the representation is closer to Neural Collapse and has higher separability.

\begin{figure*}[t]  
  \centering
  \includegraphics[width=1.0\textwidth]{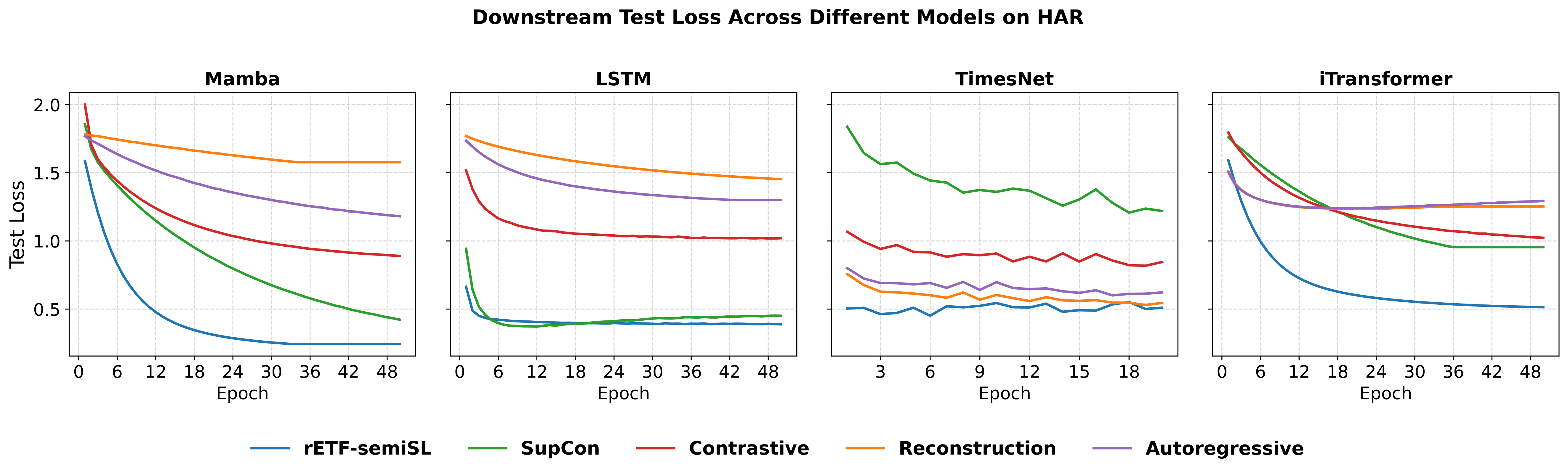}
  \caption{Cross entropy loss of downstream test set on HAR dataset. Each loss curve represents different model pretrained with different pretext task. For a certain backbone model, the fine-tune learning rates are the same.}
  \label{fig:wide_figure}
\end{figure*}

\begin{table}[t]
    \setlength{\tabcolsep}{3pt}
    \centering

    {\footnotesize
    \begin{tabular}{c|SSSS}
        \toprule
        Metric & {LogME  $\uparrow$} & {SFDA $\uparrow$} & {NLEEP $\uparrow$} & {CDNV $\downarrow$} \\
        \hline
        Auto & -0.08 & 0.40 & 0.36 & 0.13\\
        \hline
        Recon & -0.13 & 0.40 & 0.36 & 0.14\\
        \hline
        Con & -0.25 & 0.42 & 0.47 & 1.17\\
        \hline
        SupCon (30\%) & 0.47 & 0.76 & 0.74 & 0.03 \\
        \hline
        \underline{rETF-semiSL (30\%)} & \bfseries 1.27 & \bfseries 0.79 & \bfseries 0.74 & \bfseries 0.02 \\
        \bottomrule

    \end{tabular}
    }
    \caption{Transferablity Metrics and CDNV values for Pretrained Representations on HAR dataset.}\label{metrics}
\end{table}

\cref{metrics} presents the transferability metrics and CDNV values of different pretext tasks for the HAR testing set with LSTM. In order to fairly compare, we use 30\% of labelled data for Supervised Contrastive learning and rETF-semiSL. rETF-semiSL has the highest transferability value and the lowest CDNV on the three metrics, showing its advantage in balancing the label information and generalizability, and this indicates it is a better pretext task for time series classification.

%------------------------------------------------------------------

\subsubsection{Computation Complexity and Downstream Task Convergence}

Besides superior performance, rETF-semiSL is also computationally efficient compared to other pretext tasks. Given the number of classes \(C\), batch size \(N\), and embedding dimension \(d\), our pretext head exhibits a computational complexity of only \(O(N d C)\), whereas contrastive-based methods have a complexity of \(O(N^2 d)\). Moreover, unlike generative approaches, we does not require training an additional decoder, further reducing the computational overhead.
Furthermore, \cref{fig:wide_figure} shows that all models, when pretrained with rETF-semiSL, achieve a faster convergence in finetuning for downstream task.

\section{Conclusion}\label{sec:conclusion}

We propose rETF-semiSL, an ETF-based semi-supervised pretext task that explicitly enforces Neural Collapse in the encoder embeddings. This framework combines a pretext head, consisting of a fixed ETF classifier with trainable rotational matrix combined center loss, and a training algorithm, employing pseudo-labeling and gradient descent within an alternating update strategy.
We outperform on fine-tuned classification performance other pre-training methods, across multiple time series datasets and with various model backbones. Beside performance, our method also achieves the best feature transferability as well as computation efficiency and downstream convergence speed.

Additionally, we demonstrate that enforcing temporal invariance—either through our proposed forward-mixing augmentation or by combining generative tasks with our pretext task—further enhances the downstream learning capabilities of various models. We analyze the training objective of MSE-based generative tasks and show that such tasks offer limited feature separability and that their objectives fundamentally differ from Neural Collapse. Our experiment supports such claim, that using autoregressive SSL as an auxiliary task yields better performance than using it in isolation, with additional gains achieved via time series–specific augmentations such as forward mixing.

This study offers new insights into self- and semi-supervised representation learning for time series classification, highlighting the effectiveness of Neural Collapse--based pretext tasks. Our findings underscore the potential of combining discriminative and generative objectives, as well as time series--specific augmentations, to advance the state of the art in time series representation learning.

\bibliography{main}

\clearpage
\onecolumn
\appendix

\section{Appendices} \label{appendix}

\subsection{Decomposition of MSE Loss} \label{decomposition}
For a generative pretext task, we have a input signal $x \sim \mathcal{X}^t$ and a target signal $y = x^{t+k}, k\geq 0$. The model consists of an encoder $f$ and a decoder $g$, therefore We consider the prediction task where:
\begin{align}
    X &= g \circ f(x^t), X \sim \mathcal{\hat{Y}}^t  \\
    Y &= x^{t+k}, Y \sim \mathcal{Y}^t
\end{align}

The mean squared error (MSE) loss can be expressed as:
\begin{equation}
    MSE = \mathbb{E}_{\mathcal{\hat{Y}}^t, \mathcal{Y}^t} \| g \circ f(x^t) - x^{t+k} \|^2 = \mathbb{E}_{\mathcal{\hat{Y}}^t, \mathcal{Y}^t} \| X - Y \|^2.
\end{equation}

Expanding and decomposing the squared norm, we have:
\begin{equation}
\begin{aligned}
    \mathbb{E}_{\mathcal{\hat{Y}}^t, \mathcal{Y}^t} \| X - Y \|^2
    &= \mathbb{E}_{\mathcal{\hat{Y}}^t, \mathcal{Y}^t} \| X - \mathbb{E}(X) + \mathbb{E}(X) - Y \|^2 = \\
    &= \mathbb{E}_{\mathcal{\hat{Y}}^t} \| X - \mathbb{E}(X) \|^2 +
       2 \mathbb{E}_{\mathcal{\hat{Y}}^t, \mathcal{Y}^t} \big( X - \mathbb{E}(X) \big)^\top \big( \mathbb{E}(X) - Y \big)  + \mathbb{E}_{\mathcal{Y}^t} \| \mathbb{E}(X) - Y \|^2 = \\
    &=  \mathrm{Var}(X) + \|\mathbb{E}(X)\|^2 - 2 \mathbb{E}(X)^\top \mathbb{E}(Y)  + \mathbb{E}\big(\| Y \|^2\big) - 2 \mathrm{Tr}\big(\mathrm{Cov}(X, Y)\big) = \\
    &= \mathrm{Var}(X) + \|\mathbb{E}(X)\|^2 - 2 \mathbb{E}(X)^\top \mathbb{E}(Y) + \|\mathbb{E}(Y)\|^2 + \mathrm{Var}(Y)  - 2 \mathrm{Tr}\big(\mathrm{Cov}(X, Y)\big) = \\
    &= \mathrm{Var}(X) + \|\mathbb{E}(X) - \mathbb{E}(Y)\|^2 + \mathrm{Var}(Y) - 2 \mathrm{Tr}\big(\mathrm{Cov}(X, Y)\big).
\end{aligned}
\end{equation}
Finally, by substituting \( X = g \circ f(x^t) \) and \( Y = x^{t+k} \), we obtain
\begin{align}
    \mathbb{E}_{\mathcal{X}^t, \mathcal{Y}^t} \| g \circ f(x^t) - x^{t+k} \|^2
    &= \mathrm{Var}_{g \circ f}(\mathcal{X}_t) + \| \mu_{g \circ f}(\mathcal{X}_t) - \mu(\mathcal{Y}^t) \|^2 - 2 \mathrm{Tr}\big(\mathrm{Cov}(\mathcal{\hat{Y}}^t, \mathcal{Y}^t)\big) + \mathrm{Var}(\mathcal{Y}^t).
\end{align}

\subsection{Datasets Details} \label{dataset}

\textbf{HAR} The Human Activity Recognition (HAR) dataset \cite{Anguita2013APD} is built on the recordings of daily activities from 30 subjects. The record is obtained from smartphones with embedded inertial sensors that are mounted on the waist of each participant. The participants have an age range from 19 to 48 years. Each participant performs 6 activities during the recording  wearing a smartphone on the waist. The embedded accelerometer and gyroscope collect 3-axial linear acceleration and 3-axial angular velocity with a sampling rate of 50 Hz.\\
\textbf{Epilepsy} The dataset is generated by healthy participants simulating different activities including mimicking the seizure \cite{villar2016generalized}. The data was collected from 6 subjects using a tri-axial accelerometer on the dominant wrist. Each participant performed each activity 10 times at least, and the mimicked seizures were trained by a medical expert. The data was collected by a 3D accelerometer with a sampling frequency of 16 Hz.\\
\textbf{Heartbeat}
The dataset is derived from the PhysioNet/CinC Challenge 2016 \cite{liu2016open}, \cite{goldberger2000physiobank}. The heart sound recordings are collected from both healthy subjects and pathological patients from several contributors around the world. The heart sound recordings were collected from several locations on the body, including the aortic area, pulmonic area, tricuspid area and mitral area. There are two classes in the sound recordings: normal and abnormal. The normal recordings were from healthy subjects, and the abnormal ones were from patients with heart valve defects and coronary artery disease. Both healthy subjects and pathological patients include children and adults.

\subsection{Training Details} \label{appendix train}

Our training environment consists of an NVIDIA RTX3090 GPU (24GB), Intel i9-10900KF CPU, and 32GB RAM. We used PyTorch 2.4.1 with CUDA 12.4 and Python 3.10 on a Linux work station. In experiment, we set $\{123,456,789, \\101112,131415\}$ as different random seeds. In \cref{lstm details}, \cref{iTransformer details}, \cref{TimesNet details} and \cref{Mamba details}, we list important architecture and learning parameters for pre-training and linear probing of rETF-semiSL. We apply grid search on learning rates and weight decays in the given range during the pre-training stage and the linear-probing stage respectively. We also use early stop on validation set (30\% of training data) during the both stages to avoid over-fitting. 

\begin{table}[t]
\centering

% 第一行
\begin{minipage}[t]{0.48\textwidth}
\centering
\scriptsize
\caption{Details of LSTM hyperparameters}
\label{lstm details}
\resizebox{\textwidth}{!}{
\begin{tabular}{llll}
\toprule
\scshape Hyperparameters & \scshape HAR & \scshape Epilepsy & \scshape Heartbeat \\
\midrule
Recurrent Layers & 3 & 2 & 2 \\
Dimension of Embeddings & 32 & 16 & 8 \\
Max Diffusion Step & \textbackslash & \textbackslash & \textbackslash \\
Learning Rates & $[1e{-}3, 3e{-}3]$ & $[1e{-}3, 9e{-}3]$ & $[1e{-}3, 9e{-}3]$ \\
Linear Probing LR & $11e{-}4$ & $5e{-}3$ & $5e{-}3$ \\
Weight Decay & $3e{-}4$ & $3e{-}4$ & $3e{-}4$ \\
Batch Size & 32 & 128 & 32 \\
Pre-train Epochs & 50 & 100 & 150 \\
Supervised Epochs & 30 & 30 & 50 \\
$p$ & [0.01, 0.1] & [0.01, 0.1] & [0.01, 0.1] \\
$\alpha$ & 0.01 & 0.5 & 0.5 \\
Linear Probe Epochs & 80 & 150 & 150 \\
\bottomrule
\end{tabular}}
\end{minipage}%
\hfill
\begin{minipage}[t]{0.48\textwidth}
\centering
\scriptsize
\caption{Details of iTransformer hyperparameters}
\label{iTransformer details}
\resizebox{\textwidth}{!}{
\begin{tabular}{llll}
\toprule
\scshape Hyperparameters & \scshape HAR & \scshape Epilepsy & \scshape Heartbeat \\
\midrule
Attention Layers & 2 & 1 & 2 \\
Dimension of Embeddings & 32 & 8 & 32 \\
Max Diffusion Step & \textbackslash & \textbackslash & \textbackslash \\
Pre-train LR & $[1e{-}3, 3e{-}3]$ & $[1e{-}3, 9e{-}3]$ & $[1e{-}3, 9e{-}3]$ \\
Linear Probing LR & $3e{-}4$ & $11e{-}3$ & $2e{-}3$ \\
Weight Decay & $3e{-}4$ & $3e{-}4$ & $3e{-}4$ \\
Batch Size & 128 & 256 & 128 \\
Pre-train Epochs & 80 & 200 & 50 \\
Supervised Epochs & 50 & 100 & 10 \\
$p$ & [0.01, 0.1] & [0.01, 0.1] & [0.01, 0.1] \\
$\alpha$ & 0.5 & 0.5 & 0.5 \\
Linear Probe Epochs & 80 & 200 & 150 \\
\bottomrule
\end{tabular}}
\end{minipage}

\vspace{1em} % 行间距

% 第二行
\begin{minipage}[t]{0.48\textwidth}
\centering
\scriptsize
\caption{Details of TimesNet hyperparameters}
\label{TimesNet details}
\resizebox{\textwidth}{!}{
\begin{tabular}{llll}
\toprule
\scshape Hyperparameters & \scshape HAR & \scshape Epilepsy & \scshape Heartbeat \\
\midrule
Encoder Layer & 1 & 1 & 1 \\
Dimension of Embeddings & 16 & 4 & 4 \\
Max Diffusion Step & \textbackslash & \textbackslash & \textbackslash \\
Pre-train LR & $[1e{-}3, 5e{-}3]$ & $[1e{-}3, 5e{-}3]$ & $[1e{-}3, 5e{-}3]$ \\
Linear Probing LR & $3e{-}4$ & $3e{-}3$ & $2e{-}3$ \\
Weight Decay & $2e{-}4$ & $2e{-}4$ & $3e{-}4$ \\
Batch Size & 128 & 256 & 256 \\
Pre-train Epochs & 50 & 200 & 200 \\
Supervised Epochs & 30 & 100 & 50 \\
$p$ & [0.01, 0.1] & [0.01, 0.1] & [0.01, 0.1] \\
$\alpha$ & 0.1 & 0.5 & 0.5 \\
Linear Probe Epochs & 80 & 150 & 150 \\
\bottomrule
\end{tabular}}
\end{minipage}%
\hfill
\begin{minipage}[t]{0.48\textwidth}
\centering
\scriptsize
\caption{Details of Mamba hyperparameters}
\label{Mamba details}
\resizebox{\textwidth}{!}{
\begin{tabular}{llll}
\toprule
\scshape Hyperparameters & \scshape HAR & \scshape Epilepsy & \scshape Heartbeat \\
\midrule
Encoder Layer & 2 & 2 & 1 \\
Dimension of Embeddings & 32 & 32 & 32 \\
Dimension of FFN & 64 & 64 & 64 \\
Max Diffusion Step & \textbackslash & \textbackslash & \textbackslash \\
Pre-train LR & $[1e{-}3, 5e{-}3]$ & $[1e{-}3, 5e{-}3]$ & $[1e{-}3, 5e{-}3]$ \\
Linear Probing LR & $3e{-}4$ & $3e{-}3$ & $1e{-}3$ \\
Weight Decay & $0e{-}4$ & $0e{-}4$ & $0e{-}4$ \\
Batch Size & 256 & 256 & 256 \\
Pre-train Epochs & 80 & 200 & 200 \\
Supervised Epochs & 30 & 30 & 80 \\
$p$ & [0.01, 0.1] & [0.01, 0.1] & [0.01, 0.1] \\
$\alpha$ & 0.1 & 0.5 & 0.5 \\
Linear Probe Epochs & 80 & 100 & 100 \\
\bottomrule
\end{tabular}}
\end{minipage}

\end{table}

\subsection{Loss Evolution in Generative SSL Tasks}\label{sec:gen-loss-evolution}

\begin{figure}[H]
    \centering
    
    \includegraphics[width=0.5\textwidth]{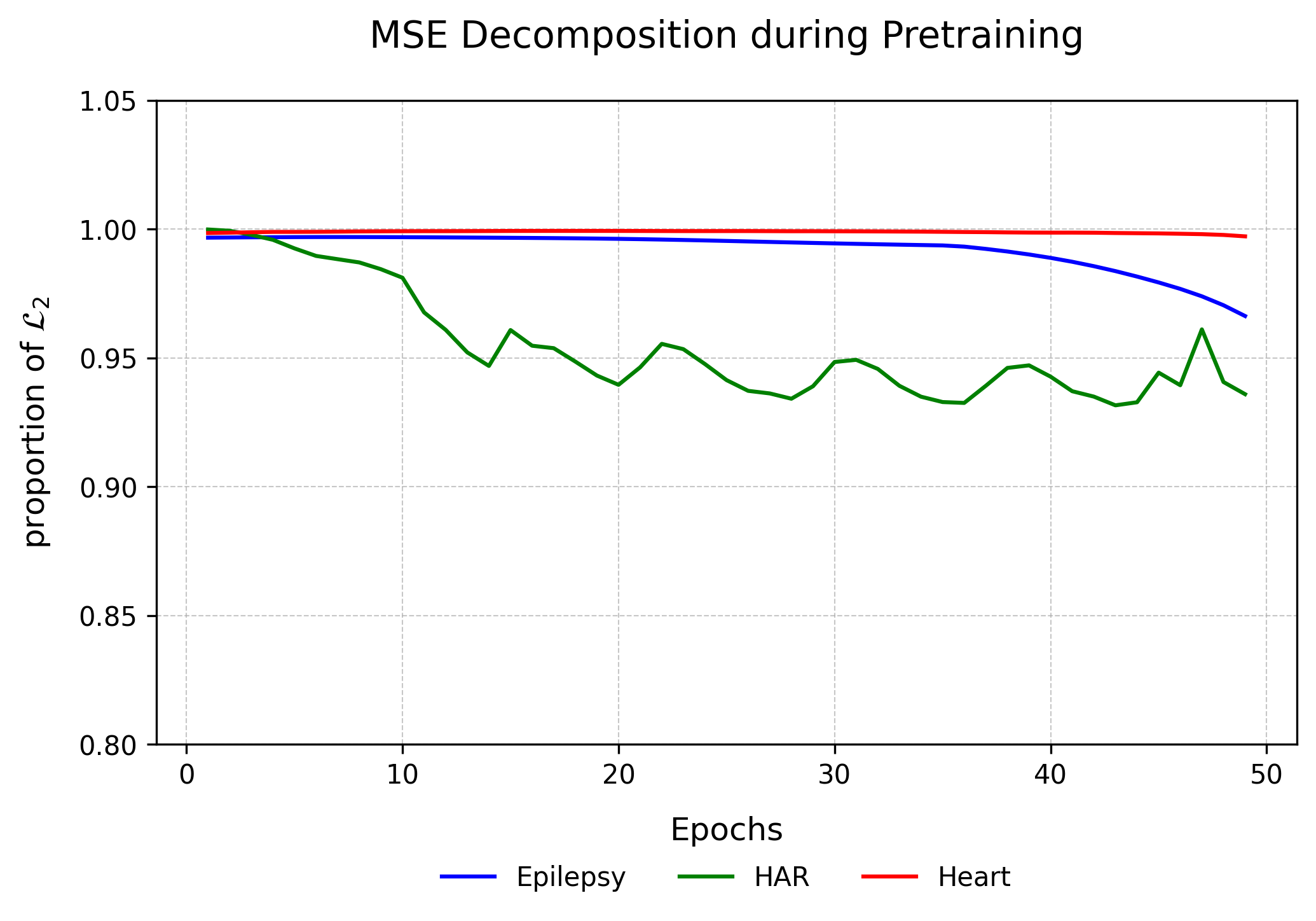}
    \caption{Relative contribution of $\mathcal{L}_2$ to the autoregressive prediction loss with LSTM, computed as $\frac{|\mathcal{L}_2|}{|\mathcal{L}_1| + |\mathcal{L}_2| + |\mathcal{L}_3|}$.}
    \label{fig:mse_decomp}
\end{figure}

To demonstrate the inherent limitations of generative SSL tasks that use MSE loss in denoising data, we decompose the overall loss based on \cref{decomposition} and analyze the contribution of the second term $\mathcal{L}_2$ during autoregressive prediction with an LSTM. Empirically, we find that $\mathcal{L}_2$ accounts for over 90\% of the total loss across all three datasets. This indicates that the model primarily focuses on aligning the mean of the reconstructed signal with that of the target signal. As a result, generative SSL with MSE loss predominantly drives the model toward matching the mean of the target distribution, while variance information is not effectively captured or optimized. Consequently, the Conditional Distribution of Noise-Variance (CDNV) in the learned embeddings remains bounded by that of the original noisy distribution.

\subsection{Experiment with Synthetic Data}\label{sec:gen-loss-evolution}

\begin{figure}[t]
    \centering
    \includegraphics[width=0.45\textwidth]{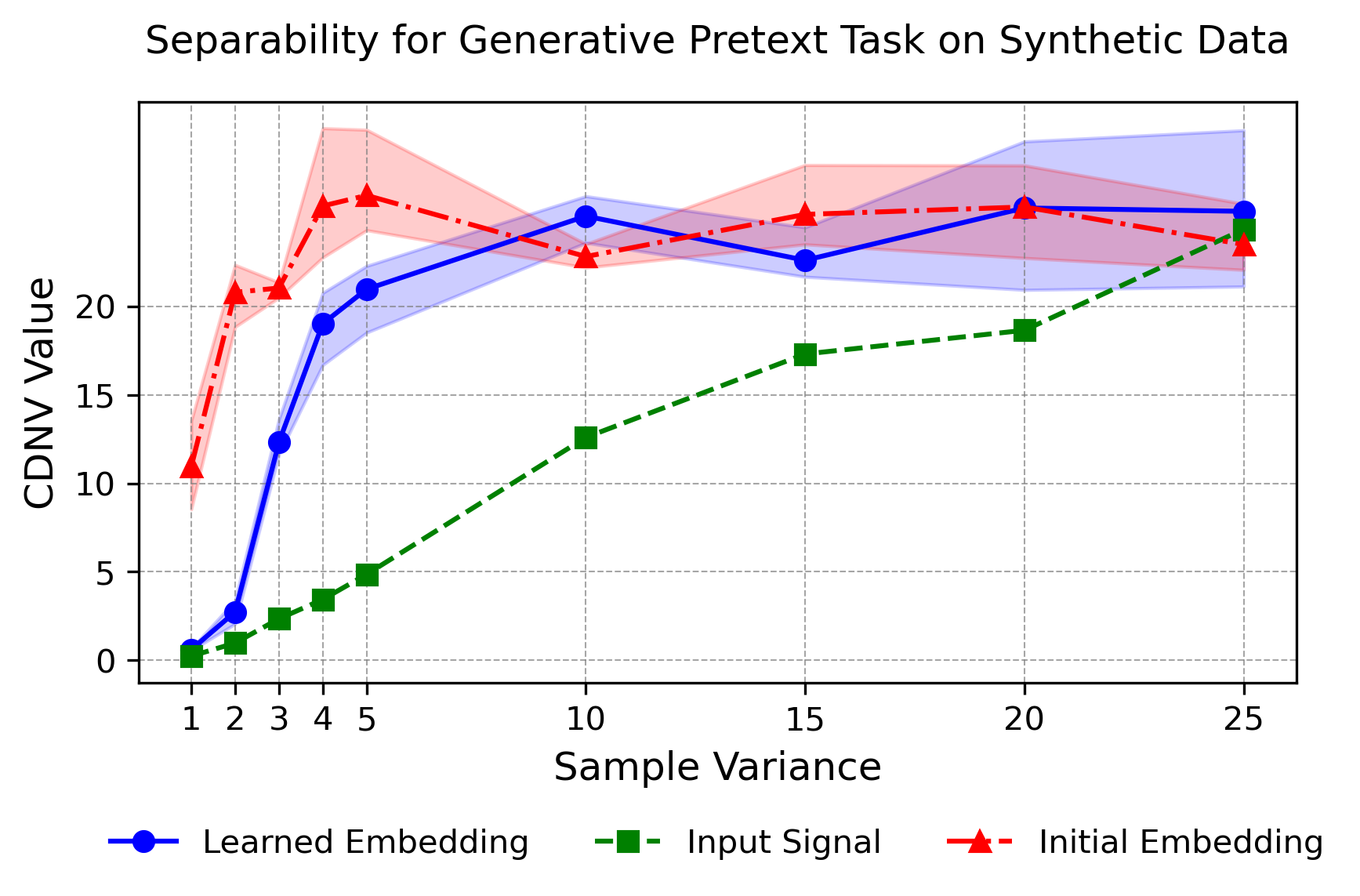}
    \caption{CDNV value of learned embeddings and input signals. Initial Embedding represents the embeddings from the initial encoder, Learned Embedding represents the embeddings from the trained encoder, and Input Signal is the discriminative subset of input data. For each variance value, we compute the CDNV with 3 random initializations of the model. The light-coloured area represents the CDNV value range, while the dark-coloured line indicates its mean value.}
    \label{fig:syn_cdnv}
\end{figure}

To further illustrate the analysis of MSE-based generative tasks, we design a synthetic experiment to empirically investigate the relationship between embedding separability and the intrinsic noise of data. Specifically, We create a synthetic classification dataset $\mathcal{D}_s$ with two classes. Each class has 1000 samples with a dimension of 400. The input data include 8 separable variables independently drawn from  Gaussian Distributions ${\mathcal{N}(-1, \sigma^2 )}$ or ${\mathcal{N}(1, \sigma^2)}$, and 392 confounding variables drawn independently from ${\mathcal{N}(0, \sigma^2)}$. By varying $\sigma^2$, we control the separability of the two classes. We perform a generative task using MSE loss and compare the CDNV of the learned embeddings against that of the input's discriminative features, across varying levels of sample variance.   

We use an MLP-based encoder-decoder structure to solve the generative task. For simplicity, we consider a reconstruction task, which means $y = x$, and we use MSE loss for optimization. The encoder $f(x) = W_2(ReLU(W_1(x)))$ projects input $x \in \mathcal{D}_s$ into an embedding vector $h \in \mathbb{R}^{8}$, which has the same size as the discriminative variables, with parameters $W_1 \in \mathbb{R}^{400 \times 32}$, and $W_2 \in \mathbb{R}^{32 \times 8}$. Similarly, the embedding vector passes through a decoder $g(h) = W_4(ReLU(W_3(h)))$ to get the output $y$, with $W_3 \in \mathbb{R}^{8 \times 32}$, and $W_4 \in \mathbb{R}^{32 \times 400}$. When calculating CDNV, we compare the embedding $h \in \mathbb{R}^{8}$ and the discriminative input variables $\mathcal{X}_d \in \mathbb{R}^8$.

As shown in \cref{fig:syn_cdnv}, the latent representations after training exhibit improved class separability compared to their initialization, yet remain bounded by the separability of the original discriminative features. As the variance increases, Neural Collapse becomes increasingly difficult to achieve through reconstruction, and the separability of embeddings approaches that of their random initialization. These results suggest that generative tasks struggle to induce Neural Collapse under high noise conditions, thereby motivating the use of augmentation-based approaches.

\end{document}